\documentclass[10pt,letterpaper]{article}

\usepackage{cogsci}
\usepackage{tipa}
\usepackage{gb4e}
\noautomath
\usepackage{graphicx}

\cogscifinalcopy 
\usepackage{inconsolata}
\usepackage{pslatex}
\usepackage[natbibapa]{apacite}
\usepackage{float} 
\usepackage[T1]{fontenc}
\usepackage[utf8]{inputenc}

\newcommand{\ententen}{\texttt{enTenTen20}}

\title{Participle-Prepended Nominals Have Lower Entropy\\Than Nominals Appended After the Participle \vspace{.25in}
}

\author{{\Large Kristie Denlinger} \quad\quad\quad\quad\quad\quad  {\Large Stephen Wechsler} \quad\quad\quad\quad\quad\quad {\Large Kyle Mahowald}\\
\texttt{\Large kdenlinger@utexas.edu} \quad \texttt{\Large wechsler@austin.utexas.edu} \quad \texttt{\Large mahowald@utexas.edu}\\ \\
         { \large Department of Linguistics}\\ { \large The University of Texas at Austin}\\ {\large Austin, TX 78712, USA}}

\begin{document}

\maketitle

\begin{abstract}
English allows for both compounds (e.g., London-made) and phrasal paraphrases (e.g., made in London). 
While these constructions have roughly the same truth-conditional meaning, we hypothesize that the compound allows less freedom to express the nature of the semantic relationship between the participle and the pre-participle nominal.
We thus predict that the pre-participle slot is more constrained than the equivalent position in the phrasal construction. 
We test this prediction in a large corpus by measuring the entropy of corresponding nominal slots, conditional on the participle used.
That is, we compare the entropy of $\alpha$ in compound construction slots like ``$\alpha$-[V]ed'' to the entropy of $\alpha$ in phrasal constructions like ``[V]ed by $\alpha$'' for a given verb V.
As predicted, there is significantly lower entropy in the compound construction than in the phrasal construction.
We consider how these predictions follow from more general grammatical properties and processing factors.

\textbf{Keywords:} 
compounding; word-formation; entropy; participles; information-theoretic linguistics; corpus analysis
\end{abstract}

\section{Introduction}
Compound participles (e.g. \textit{women-led}) often have phrasal paraphrases (e.g. \textit{led by women}).  The two forms are truth-conditionally equivalent, and yet they differ in usage  \citep[see, e.g.,][for related discussion of German adjectival participles]{Gehrke2017,Maienborn2009,Maienborn2016}. For instance, consider the following:

\begin{exe}
\ex \label{Chomsky} \begin{xlist}
\ex a Wright-designed house
\ex a house designed by Wright
\end{xlist}
\ex \label{Charlotte}\begin{xlist}
\ex ?a Charlotte-designed house
\ex a house designed by Charlotte
\end{xlist}
\end{exe}
 Unlike the phrasal reduced relative construction `designed by $\alpha$' (\ref{Chomsky}b, \ref{Charlotte}b), the compound construction `$\alpha$-designed' (\ref{Chomsky}a,  \ref{Charlotte}a) is constrained to uses whereby the participial modifier $\alpha$ is an identifiable name. More broadly, it seems that the compound participle favors well-established concepts, making  \textit{doctor-prescribed} and \textit{earthquake-destroyed} sound better than (\ref{doctordestroy}a):  

\begin{exe}
        \ex \label{doctordestroy}\begin{xlist}
         \ex ?doctor-destroyed files
        \ex files destroyed by a doctor
    \end{xlist}
\end{exe}
In contrast to the compounds, all of the corresponding phrasal constructions are equally felicitous: \textit{prescribed by a doctor, destroyed by an earthquake}, or (\ref{doctordestroy}b).

\begin{figure}[t]
   \centering
    \includegraphics[width=\columnwidth]{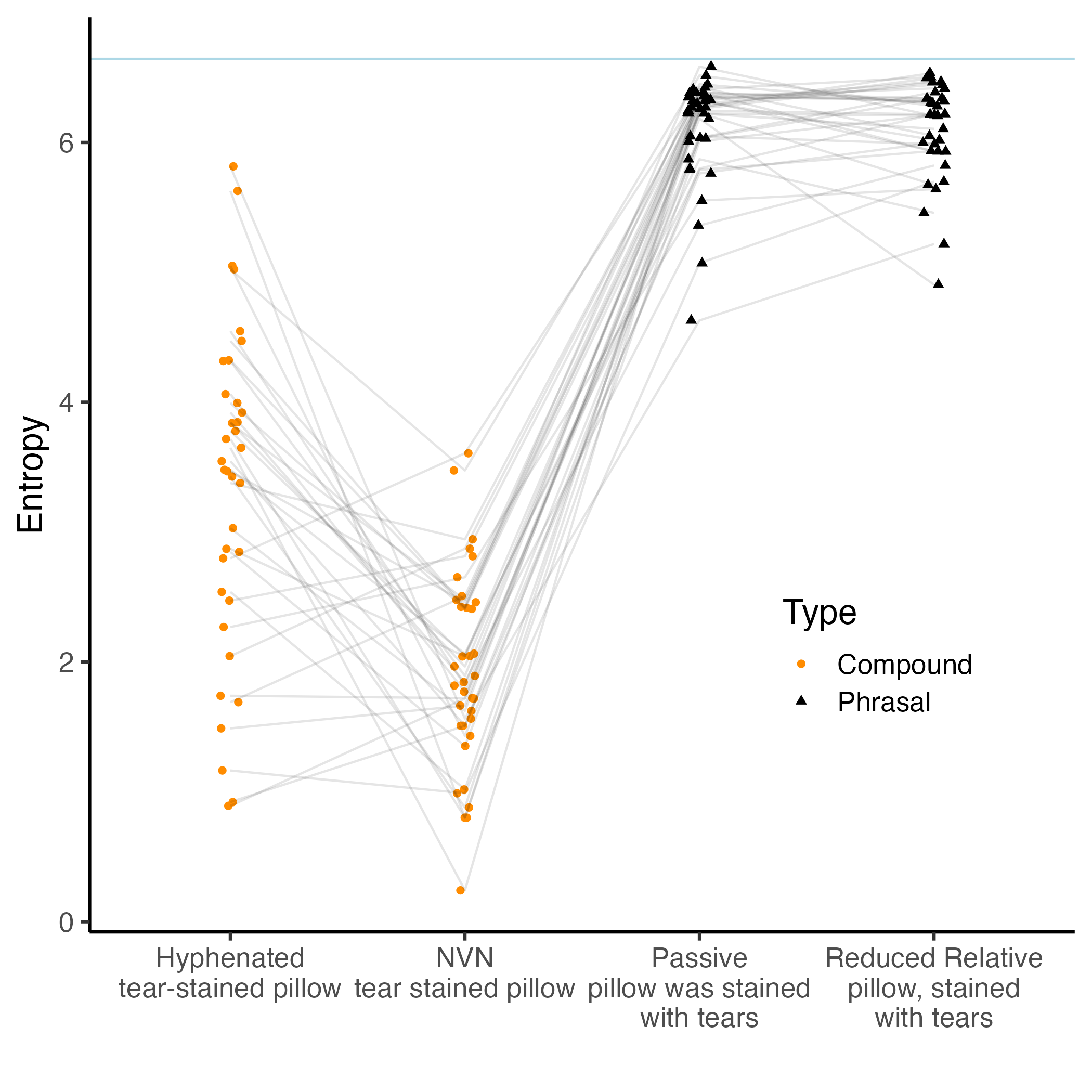}
    \caption{Split by construction, the entropy over elements $\alpha$ for each participle. Passives and reduced relatives (the phrasal constructions) show consistently higher entropy than the Hyphenated compound and unhyphenated (Noun Verb Noun) compound constructions. The maximum possible entropy is indicated by the blue horizontal lines in the figures below.}
    \label{fig:Entropy}
\end{figure}

The patterns displayed in (\ref{Chomsky}-\ref{doctordestroy}) reflect preferences rather than strict rules. Given a rich enough context, speakers \textit{can} refer to a property as `Charlotte-designed' or `doctor-destroyed,' and listeners will use an array of techniques to interpret the compound meaning, such as analogy to similar established meanings in their lexicon \citep[see, e.g.,][]{Gagne2002,Libben2020,Stekauer2012}. In other words, compounding is a compositional strategy that in principle can express any number of relations between constituents, yet in practice tends to be reserved for expressing meanings that speakers find name-worthy and listeners find predictable. 
From the construction grammar perspective, compounds and phrases are both constructions which are partially productive to different extents \citep{goldberg2019explain}.

If speakers are more conservative when choosing to compose meanings with compounds, we expect to see less variability (lower entropy) in the constituents that are used in such constructions compared to those that are used in comparable phrasal constructions. 
In this study, we examine the extent to which this prediction is borne out in a large corpus.  

This hypothesis is motivated by three factors involving different aspects of the grammatical form of the compound, relative to the phrase. 
We do not attempt to adjudicate between these factors here; for our purposes, they make similar predictions.
First, word structure emerges through processes of phonological simplification that are fed by higher frequency, leading to morphologization for the more predictable combinations \citep{Bybee2000}.  Second, the prenominal position is consistent with a well-established `easy-first' preference: more predictable elements come first in a sequence \citep{macdonald2013language}.  Third, the loss of the preposition in the compound, relative to its presence in the phrase, leads to a decrease in expressivity.  This is compensated by the lower information content of predictable combinations \citep{Shannon1948}.

In the study described below, we compared the variability of constituents found in participial compounds (e.g. $\alpha$-made) to those found in corresponding phrasal constructions (e.g. made by $\alpha$). We did this by measuring the entropy of compound and phrasal constructions in a corpus of English internet text, \ententen, conditioned on the participle.
We focused on two types of compound constructions (hyphenated and unhyphenated 
compounds) and two types of phrasal constructions (passives and reduced relatives).
We predict that, conditional on the participle, compounds will have lower entropy than their phrasal paraphrases. That is, conditional on a given participle P, the $\alpha$ element in $\alpha$-P will be on average less suprising than the same $\alpha$, conditional on P, used in a phrasal construction.

To foreshadow our results: We found that entropy was significantly lower in compound constructions than in equivalent phrasal constructions using the same participle.  Specifically, given a particular participle form (\textit{stained}), the noun appearing with it (\textit{tear}) is significantly more predictable in the compound construction (\textit{tear-stained}) than in the phrasal paraphrase (\textit{stained with tears}).  This confirms the intuition that the compounds are more well-established combinations than the phrases.

\section{Methods}

The general methodology for this study is to first identify a set of participles that can occur in compound constructions---specifically the hyphenated construction as in \textit{tear-stained pillow} or the unhyphenated noun-verb-noun (henceforth NVN) construction as in \textit{tear stained pillow} without the hyphen, as well as in phrasal constructions---specifically the passive \textit{pillow was stained with tears} and reduced relative \textit{pillow stained with tears}.
We then extract several hundred examples of a given participle $P$ used in each construction from a corpus, parsing them using \texttt{spaCy} to identify the relevant nominal $\alpha$.
We then downsample so we have 100 examples of each construction and compute the entropy, which allows us to test our key hypothesis.
We detail these steps further in the following sections.

\subsection{Data}

This study uses The English Web Corpus, \ententen, which belongs to the TenTen corpus family developed by Sketch Engine \citep{jakubivcek2013}. The 2020 iteration of the corpus used in this study,  \ententen{} contains about 36 billion words and includes a range of web-based texts across a number of genres.\footnote{Top genres include Culture and Entertainment, Travel and Tourism, Religion, Technology and IT, Economy Finance and Business, and Science.} 

We choose this corpus in part because of its large size: even the most frequent participles used in compounds are relatively rare in a corpus overall. For example, in  \ententen{} there are 5,981,951 tokens of compound participles with the participle \textit{based} (e.g. \textit{community-based}, \textit{evidence-based}, \textit{text-based}) and yet these make up only 0.017$\%$ of the corpus. The less common, albeit still familiar, participle  \textit{stained} (e.g. \textit{blood-stained}, \textit{tear-stained}, \textit{grass-stained}), appears in 30,526 tokens comprising 0.00000847$\%$ of the corpus. The even rarer participle \textit{negotiated}, (e.g. \textit{peer-negotiated}, \textit{union-negotiated}, \textit{market-negotiated}) appears in only 5,493 tokens, comprising 0.000000152$\%$ of the corpus.

\ententen{} is designed to pull only linguistically valuable content and discards between 40-60$\%$ of downloaded text to avoid spam, machine-generated and otherwise unwanted content so that it can be used as a valid proxy for language use. Additionally, the corpus undergoes a deduplication process which removes identical or slightly altered versions of the same content at the paragraph level. Still, with such a large corpus there is always some unwanted content. So tokens that do not seem to represent natural language use, for instance advertisements with incomplete sentences, were excluded manually when possible.

The corpus is lemmatized and tagged with a Penn Treebank dataset for part of speech. These tags include a specific tag for participles. Simple lemmas were searched using the Sketch Engine web interface and more complex grammatical structures were queried using a special corpus query language (CQL). The majority of data used in this study was pulled from the corpus using CQL.

\subsection{Identifying Participles for Study}

To generate a list of participles to use as a test set, we began by pulling a randomized sample of all verbs which are tagged as participles in the corpus. We selected the first participles with sufficient usage in all of the constructions under inquiry. There are some participles which occur in the relevant phrasal constructions but are not found in compounds, or else are found quite rarely. For example, the verb \textit{tussle} only has 13 tokens in the corpus in the form $\alpha$-tussle (e.g. \textit{wind-tussled hair}). There are only 2 tokens of the form $\alpha$-recoil (e.g. \textit{short-recoiled barrel}). Compounds of the form $\alpha$-\textit{eavesdropped} don't appear in the corpus at all. We pulled a random group of participles that had at least 200 tokens in each of the constructions below. 

This pass led to 65 participles in total, although only 36 produced enough valid parses for analysis (as described in the section on Obtaining Parses). See the x-axis in Figure \ref{fig:Entropy2} for the complete list of participles ultimately used for analysis. Although this strategy biases our sample towards more frequently used participles, this method ensures that we have sufficient examples of both compound and phrasal constructions for a given participle to calculate a comparable measure of entropy.

\subsection{Pulling Relevant Corpus Data}

After identifying the relevant participles, we pulled data for them in each of 4 constructions (2 phrasal and 2 compound), as described below\footnote{Code and Datasets available at: github.com/kdenlin/Entropy}.

\subsubsection{Phrases}
We pulled data for participles used in two phrasal constructions: passives and reduced relative clauses. For instance, in (\ref{DCs}) the participle is \textit{stained}, the head noun is \textit{the pillow} and the $\alpha$ is \textit{tears}.

\begin{exe}
    \ex \label{DCs}\begin{xlist}
        \ex \textsc{Passive}: \\ \textit{The pillow} was \underline{stained} by/with \textbf{tears}
        \ex \textsc{Reduced Relative}: \\\textit{The pillow} \underline{stained} by/with \textbf{tears}
    \end{xlist}
\end{exe}

We want to pull phrasal paraphrases that can be directly compared to compounds. In the examples in (\ref{DCs}), \textit{tears} plays an instrumental participant role in the staining event and is introduced by a prepositional phrase. The $\alpha$s in participial compounds can also be adverbs and adjectives in addition to nominal participants (e.g. \textit{disgustingly-stained}, \textit{half-stained}). The corresponding adjectives and adverbs surface in many places in phrasal constructions, so it was best to limit our search to only phrases in which a nominal $\alpha$ is introduced via an immediately preceding participle P, such as the examples above, and compare those to uses of nominal $\alpha$s in compounds. In other words, \textit{$\alpha$-P} always matched with \textit{P-Prep-$\alpha$} for all values of $\alpha$.  

Queries for these constructions are listed below, in CQL followed by an explanation and an example for each. Brackets represent word boundaries and tags represent part of speech classifications assigned to lemmas via the Penn Treebank dataset. 

\begin{exe}
    \ex \textsc{Passive}:
    \begin{xlist}
        \ex $[$tag = ``VB.*"$]$  $[$tag =``RB"$]$? $[$tag = ``VVN" $\&$ lemma = ``stain"$]$  $[$tag=``IN"$]$ within $<$ s$/>$ 
        \ex $[$ be verb ] $[$optional adverb] $[$ participle with the lemma `stain'$]$ $[$preposition$]$ within a sentence
        \ex $[$is$]$ ($[$very$]$) $[$stained$]$ $[$with$]$ 
    \end{xlist}
\end{exe}
 For the passive query, the head noun is the lemma immediately to the left and the $\alpha$ is immediately to the right. Therefore, we filtered out cases where the passive was embedded in a relative clause (e.g. \textit{the pillow which was stained with tears}) by removing tokens where the lemma to the left was \textit{which} or \textit{that}. 

\begin{exe}
    \ex \textsc{Reduced Relative}:
    \begin{xlist}
        \ex $[$tag = ``N.*" $\&$ tag!=``NNSZ" $\&$ tag!=``NNZ" $\&$ tag!=``NPZ" $\&$ tag!=``NPSZ"$]$  $[$tag = ``VVN" $\&$ lemma = ``stain"$]$ $[$tag=``IN"$]$ within $<$ s$/>$ 
        \ex $[$ (Non-Possessive) Noun $]$ $[$ participle with the lemma `stain'$]$ $[$preposition$]$ within a sentence
        \ex $[$pillow$]$ $[$stained$]$ $[$with$]$ 
    \end{xlist}
\end{exe}
For this query, the head noun is the first lemma in the query and the $\alpha$ is the first noun to the right. We did not allow for optional adverbs in this construction as in the passives because it significantly slowed down processing time.

\subsubsection{Compounds}
In order to pull tokens of participles being used in compounds, we want to look for cases where the $\alpha$ immediately precedes the participle as in the following:

\begin{exe}
\ex \label{panda}\begin{xlist}
    \ex the \textbf{tear}(-)stained \textit{pillow}
    \ex the \textit{pillow} is \textbf{tear}(-)stained
    \end{xlist}
    \end{exe}
In the present study, we focus on pulling prenominal participles, as in (\ref{panda}a), with nominal $\alpha$s since it is easier to identify these strings using the Sketch Engine query language. There are a few ways to pull these tokens, each with their own positives and negatives. The first is to look for $\alpha$s adjacent to prenominal participles. We'll call this the NVN construction. We can pull this using the following query: 

\begin{exe}
    \ex \textsc{NVN Construction}:
    \begin{xlist}
        \ex $[$tag = ``N.*" $\&$ tag!=``NNSZ" $\&$ tag!=``NNZ" $\&$ tag!=``NPZ" $\&$ tag!=``NPSZ"$]$ $[$tag = ``VVN" $\&$ lemma = ``stain"$]$ $[$tag = ``N.*"$]$ within $<$ s$/>$ 
        \ex  $[$ (Non-Possessive) Noun $]$ $[$ participle with the lemma `stain'$]$ $[$ Noun $]$
        \ex $[$ tear $]$ $[$ stained $]$ $[$ pillow $]$
    \end{xlist}
\end{exe}
While this query does pull certain participial compounds, it also pulls a lot of other data that is hard to control for. For example this query gives us tokens where $\alpha$ heads a relative clause (e.g. \textit{one reason stained glass became popular}), where it modifies the head independently of the participle (e.g. \textit{commemorative stained glass window}) 
or where it plays another role such as the direct object of a ditransitive verb (e.g. \textit{I taught adults stained glass techniques}). 

In these examples the bigrams \textit{reason stained}, \textit{commemorative stained}, and \textit{adults stained} are not compounds. So, this query results in a lot of noise. 
We control this noise by obtaining parses, as described in the next section, and we evaluate these constructions separately from the more easily regex-ed and parsed hyphenated constructions.

We also pull hyphenated participles. The asterisk in the query indicates an absent element, so this searches for all words in which $\alpha$ precedes a hyphen and the participle in question. 

\begin{exe}
    \ex \textsc{Hyphenated Compound}
    \begin{xlist}
        \ex $[$*-stained$]$
        \ex tear-stained
    \end{xlist}
\end{exe}
 Results from the query above were filtered for hyphenated compounds that precede a noun. 
There is a lot less noise in this query than the NVN construction.  However, the downside is that hyphenated compounds are lemmatized as one word, preventing us from controlling for the part of speech of stems within words. This means that we cannot ensure that the $\alpha$ is a noun, which is crucial to make a fair comparison to phrasal constructions since we identify only phrases with a nominal $\alpha$ introduced by a prepositional phrase. For both of the compound queries, we processed the raw data through a parser, to obtain data on the relevant participles and $\alpha$s.

For each participle, up to 5,000 tokens were pulled in each of the four constructions, totaling 1,144,828 total sentences containing participles in one of the four constructions above. 

\subsection{Obtaining Parses}

To compute our key entropy measures, we need to know which elements of a sentence correspond to the $\alpha$ element and the participle. To do this, we obtain dependency parses using spaCy. Depending on the construction, the $\alpha$ has a different dependency relationship with the participle and the head noun. For compounds, the $\alpha$ will modify the participle, which will modify the head noun. For passives and reduced relatives, the $\alpha$ will head a NP within a PP which is a dependent of the participle.

In addition to identifying the $\alpha$ and participle elements, passing the data through the parser allows us to filter out noisy data from the dataset, that is, data that didn't actually pull the participle used in the construction intended by the query. As discussed above, while the CQL queries are generally able to identify tokens of participles in the desired constructions, they also tend to pull tokens of participles used in constructions with similar surface order patterns but the wrong dependency relationships. If a participle ended up with less than 100 tokens in a construction after being parsed, it was excluded completely. Otherwise, for each participle, we randomly sampled 100 tokens in each construction. With this filter, we ended up testing 36 participles across the 4 constructions\footnote{A full list of the participles tested can be found on the x-axis of Figure \ref{fig:Entropy2}.}. For these participles, we then extracted lists of $\alpha$s in each construction to calculate entropy.

\subsection{Measuring Entropy} 

To measure uncertainty, we use conditional Shannon entropy to calculate surprisal in bits \citep{Shannon1948}.
Specifically, we compute the conditional entropy of the $\alpha$ element, given the participle $P$:

\begin{exe}
\ex $H(\alpha|P) = -\sum_{i} p(\alpha_i|P) log_2 p(\alpha_i|P)$
\end{exe}

\noindent 
Higher Shannon entropy corresponds to \textit{more uncertainty} about the identity of $\alpha$ given the participle, and lower Shannon entropy corresponds to lower uncertainty about $\alpha$ given the participle.
Intuitively, if every time the participle \textit{stained} occurred in a compound, it occurred with the $\alpha$ \textit{tear} to form \textit{tear-stained}, our conditional entropy measure would be low (0 in that case).
If every time the participle \textit{stained} occurred in a phrasal construction, it occurred with a different unique $\alpha$ (e.g., \textit{stained with ink}, \textit{stained with yogurt}, \textit{stained with ketchup}, etc.), the entropy would be high.
We obtain a measure of entropy for each participle, in each of the relevant constructions (NVN, hyphenated compounds, passives, reduced relatives).
The prediction is that hyphenated compounds and NVN's will have lower entropy than passives and reduced relatives.

\section{Results}

\begin{figure*}[t]
   \centering
    \includegraphics[width=.9\textwidth]{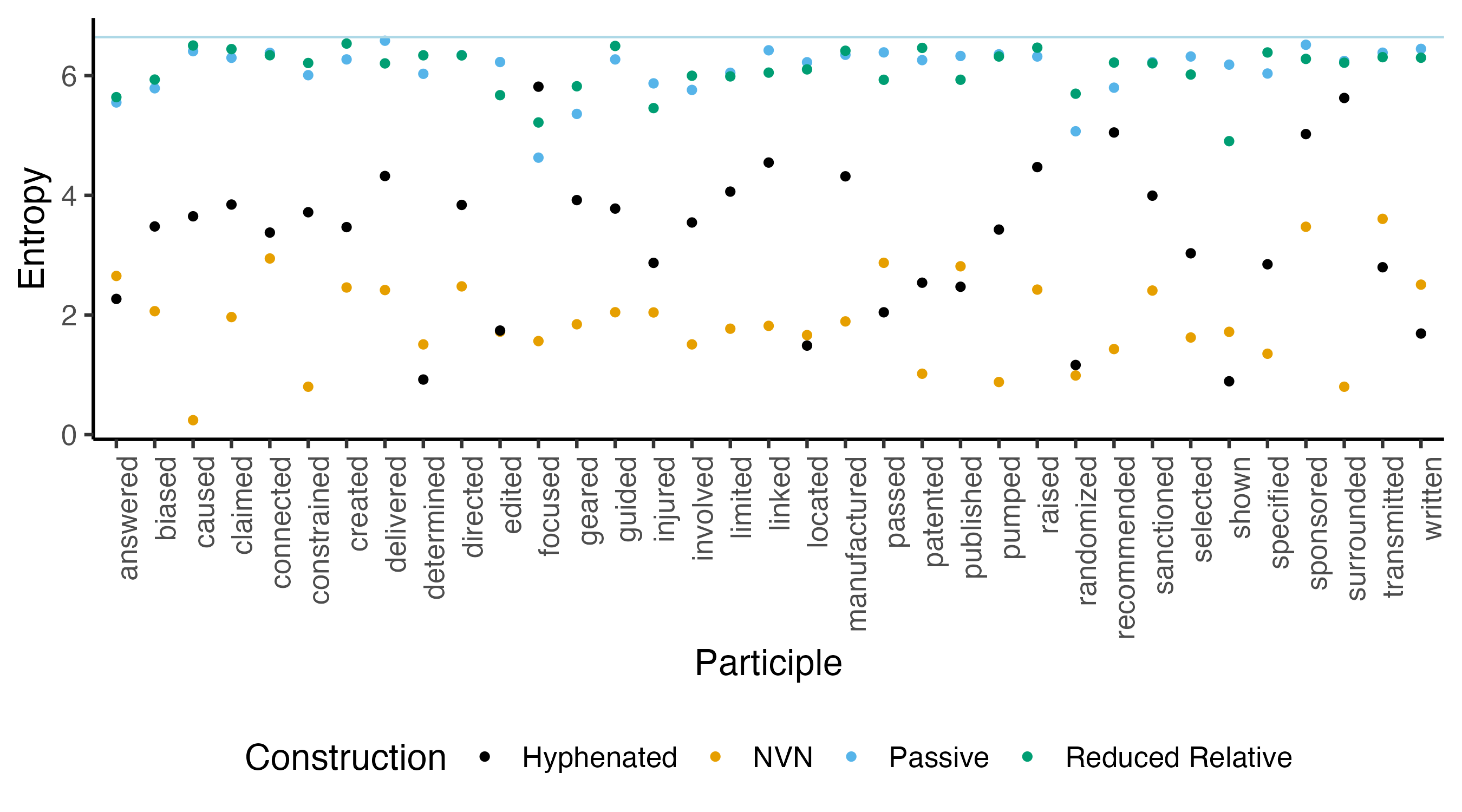}
    \caption{Split by participle, the entropy over 100 occurrences for each construction. There is considerable variability by participle, but in general the compound constructions show lower entropy than the phrasal constructions. Since entropy was measured across 100 random tokens, the maximum entropy measure is 6.64 (i.e. -log2(1/100)). This would be the case where all 100 $\alpha$s are distinct. The maximum entropy measure is indicated by the blue horizontal 
    line at the top of the diagram.}

    \label{fig:Entropy2}
\end{figure*} 

For each participle, we calculated a measure of $\alpha$ entropy across 100 random tokens for each verb of the four different constructions. 
Because the number of tokens is held constant across constructions, entropies are directly comparable.
These results are illustrated in Figure \ref{fig:Entropy} and Figure \ref{fig:Entropy2}. In Figure \ref{fig:Entropy}, the four constructions are represented along the x-axis and entropy measures for each verb across these constructions are plotted on the y-axis. Both passive and reduced relative uses have higher entropy overall. Additionally, both compound constructions have a noticeably wider range of entropy measures across individual participles than the phrasal constructions, especially the hyphenated compounds. In Figure \ref{fig:Entropy2}, each verb is plotted on the x-axis and the constructions are color-coded. The passive and reduced relative entropy measures are both quite high, while the NVN and hyphenated entropy measures are lower and cover a wider range.

To assess significance, we ran a mixed effect model using the \texttt{lme4} package \citep{bates2014fitting}, predicting the entropy from the construction (treating the Hyphenated construction as the baseline), with a random intercept for the Participle. 
This constitutes the maximal random effect structure, as recommended by \citet{barr2013random}.
Comparing the full model to an otherwise identical model without the fixed effect of construction using a likelihood ratio test, we found that including construction information significantly improved the model ($\chi^2(3) = 260.79$, $p < .0001$).
Inspecting the fixed effect coefficients, we found strong evidence for our predictions. The Passive ($t=15.33$, $p < .0001$) and Reduced Relative ($t=15.33$, $p < .0001$) constructions both showed significantly higher entropy than the Hyphenated construction.

We found that NVN construction showed significantly lower entropy than the Hyphenated construction ($t = -7.65$, $p < .0001$).
Further work is needed to assess the source of this difference, although it is relatively small compared to the difference between phrasal and compound constructions more generally.
Crucially, it is clear that both hyphenated constructions and NVN constructions show less entropy than either phrasal construction, and so the main result is not dependent on the presence of the hyphen.

Running a new model with just the phrasal constructions (Passive and Reduced Relative) included and comparing to a model without construction as a predictor, we found that there was no significant difference between Passives and Reduced Relatives ($\chi^2(1) = 0.0247$, $p = .88$).
For some participles in these categories, most of the $\alpha$s were unique, and so there is some possibility of a ceiling effect which makes it hard to observe differences between passives and reduced relatives. 
Future work could explore this in a larger sample.

\section{Discussion}
 The results of this study support the hypothesis that $\alpha$s used in compounds have lower entropy than those used in phrasal paraphrases. So, 
 hearers are relatively less surprised about
 $\alpha$ in a compound than in a phrase. Additionally, the entropy measures in compound constructions were more varied across all participles than those of the passive and reduced relative constructions.  
As noted in the introduction, we see three ways in which differences in the form of the compound and the phrase predict these differences in entropy.

 Previous research has shown that internal constituents of a word are more predictable than the constituents of a phrase \citep{mansfield2021}.  Here, the notions \textsc{word} and \textsc{phrase} are grammatical categories distinguished by some general cross-linguistic tendencies but also language-specific structural characteristics. For example, English phrases are left-headed (\textit{song of a bird}) while words are right-headed (\textit{birdsong}).    One possible explanation for the greater within-word predictability is that word structure itself emerged through processes of phonological simplification that affect language change by amplifying slight biases over many iterations.  More frequent collocations show a higher rate of change than less frequent ones, leading to morphologization for the more predictable ones \citep{Bybee2000,phillips1984}. However, the noun-participle compound word structure was established long ago. If a current speaker is still more apt to choose the compound over the phrase when uttering a more predictable combination of words, then one possibility is that speakers have intuitions of predictability, know that word structures are appropriate for high-predictability collocations, and tailor their utterances accordingly. 
 
 The idea that speakers use words and phrases for different communicative purposes is far from novel. 
 
 \cite{Downing1977} suggests that word-formation processes are used to name, while syntactic composition is used to assert or describe. A crucial difference between these goals is that names tend to be consistent across contexts, while descriptions and assertions are highly context-specific. For our purposes, the fact that speakers use compounds to name a subset of concepts suggests that compounding is specialized for meanings that tend to be reusable, a factor present in more recent models of derivational morphology \citep{ODonnell2011a}.

A second aspect of the compounds' form leading us to expect greater predictability is their position to the left of the noun that they modify.  
There is a well-established preference for ``easy'' or more predictable elements to come first in a sequence \citep{macdonald2013language}.
For instance, pre-nominal gender markers reduce uncertainty about the upcoming noun \citep{dye2018alternative}, as can prenominal adjectives in languages where adjectives precede nouns \citep{dye2017cute}.
A related prediction is that elements that occur before the participle are more likely to be ones that reduce uncertainty about the upcoming participle.

Finally, the extent to which we observe lower entropy in a construction is likely influenced by the expressiveness of the construction itself. The phrases used in this study differ from the compounds in that they contain a preposition, which can help facilitate their interpretation. For instance, the preposition \textit{in} aids the interpretation that $\alpha$ is a location in \textit{made in London}. The constituents of compounding are only related formally by virtue of their juxtaposition, and interlocutors must rely to some extent on idiosyncratic lexical knowledge to infer an appropriate relation. The semantic relationships between two elements in a compound vary widely across particular lexical items. For instance, in compounds of the form $\alpha$-made, $\alpha$ can play the role of an agent (e.g. \textit{artisan-made}), a location (e.g. \textit{London-made}), an instrument (e.g. \textit{hand-made}), or a goal (e.g. \textit{purpose-made}) in the underlying making event. 

The presence of a preposition in phrasal constructions doesn't mean that their interpretation doesn't require any idiosyncratic knowledge at all. It has been shown that English prepositions are inherently polysemous \citep[see, e.g.,][\textit{inter alia}]{Kreitzer1997,tyler2007,Rice1992}. A reduced relative clause containing the preposition \textit{by} may also introduce an $\alpha$ with a similar range of roles as those found in compounds, such as an agent (e.g. \textit{research conducted by students}), a cause (e.g. \textit{person bound by a spell}), an instrument (e.g. \textit{vase sculpted by hand}) or some other more idiosyncratic role (e.g. \textit{students sorted by height}). Therefore, the interpretation of both compounds and phrases undoubtedly relies to some degree on idiosyncratic lexical knowledge to different degrees. Crucially, there seems to be a trade-off between 
expressivity and predictability such that if the construction has more functional elements aiding interpretation, then speakers can use that construction to express more surprising information.

The hypothesis in this study could reasonably be extended to other types of word-like and periphrastic constructions. For instance, we might hypothesize that noun-noun compounds, such as \textit{girlfriend} have lower entropy relationships than approximate periphrastic paraphrases, such as \textit{a friend who is a girl}. However, the case explored here capitalizes on the fact that there are relatively consistent constructions for composing phrases with adjectival participles. In contrast, periphrases for noun-noun compounds are quite variable so would be difficult to quantify. 

  The fact that there was a statistically significant difference between hyphenated participles and NVN constructions (i.e. \textit{tear-stained} vs. \textit{tear stained}) raises additional questions regarding what other mitigating factors may govern the entropy of participles in a written corpus. Perhaps entropy is an implicit aspect of a speaker's linguistic competence that influences their decision to use a hyphen while writing out a compound. Or perhaps there is some other attribute in the data affecting the entropy values. This is a potential question to be explored in future work. 
  
  Additionally, the sample of verbs used in this study was limited to only those with sufficient frequency across all four relevant constructions. With adapted methodologies, future work may examine whether these findings generalize to more infrequent verbs, which may have different sensitivities to the entropy effect. We could also extend the present study to a wider and/or more fine-grained breakdown of construction types in English. For instance, we would expect participles used in perfects to pattern with other phrasal constructions. Finally, we focus here on English compounds but the motivations we discuss are in principle universal information processing factors. Therefore, we expect that similar quantitative differences between word-like and phrase-like constructions may be expected in other languages.

\section{Conclusion}

In this study, we identify a difference in the use of compound and verbal constructions by examining patterns in corpus data. Quantitative approaches afford us the ability to explain certain facts about grammar that can't be handled with ontological semantic category distinctions. A compound like \textit{tear-stained} and phrase like \textit{stained with tears} have the same meaning and logical type, yet our results suggest that the prenominal position seems to be specialized in English for only uses of participles as names for properties. Therefore, the distinction between compounding and syntactic composition is not only a strategy for speakers to derive new words, but also a difference recognized by the grammar. The findings in this study support other work connecting linguistic principles to usage patterns in large corpora, an approach not only beneficial for those interested in theoretical linguistics but also for those interested in bridging the gap between human language processing and computational large language models.

\section{Acknowledgments}
We would like to thank Hans Kamp, David Beaver, John Beavers, Louise McNally and members of the UT-Austin Syntax and Semantics seminar group for valuable feedback and discussion. We would also like to thank our anonymous CogSci reviewers for their helpful comments.

\bibliographystyle{apacite.bst}

\setlength{\bibleftmargin}{.125in}
\setlength{\bibindent}{-\bibleftmargin}

\bibliography{library}

\end{document}